\documentclass{article}
\usepackage{spconf,amsmath,epsfig}
\usepackage{pdfpages}
\usepackage{CJK}
\usepackage{booktabs}
\usepackage{makecell}
\usepackage{amssymb}
\usepackage{graphicx}
\usepackage{tabularx}
\usepackage{array}
\usepackage{algorithm}
\usepackage{algpseudocode}  
\usepackage{cite}  
\usepackage{threeparttable}
\usepackage{color}
\usepackage{bbding}
\usepackage{multirow}
\usepackage{verbatim}
\usepackage{url}
\usepackage[colorlinks = true,
            anchorcolor = blue]{hyperref}
\title{RSSOD-Bench: A large-scale benchmark dataset for Salient Object Detection in Optical Remote Sensing Imagery}
\name{Zhitong~Xiong$^{1}$, Yanfeng~Liu$^{2,3}$, Qi~Wang$^{3}$, Xiao Xiang Zhu$^{1}$}
\address{$^{1}$  Data Science in Earth Observation, Technical University of Munich (TUM), Ottobrunn, Germany\\
$^{2}$ School of Computer Science, and $^{3}$ School of Artificial Intelligence, Optics and Electronics (iOPEN), \\Northwestern Polytechnical University, Xi’an 710072, Shaanxi, P. R. China\\
}

\begin{document}
%\ninept

\maketitle

\begin{abstract}
We present the RSSOD-Bench dataset for salient object detection (SOD) in optical remote sensing imagery. While SOD has achieved success in natural scene images with deep learning, research in SOD for remote sensing imagery (RSSOD) is still in its early stages. Existing RSSOD datasets have limitations in terms of scale, and scene categories, which make them misaligned with real-world applications. To address these shortcomings, we construct the RSSOD-Bench dataset, which contains images from four different cities in the USA \footnote{The RSSOD-Bench dataset can be accessed via \url{https://github.com/EarthNets/Dataset4EO}}. The dataset provides annotations for various salient object categories, such as buildings, lakes, rivers, highways, bridges, aircraft, ships, athletic fields, and more. The salient objects in RSSOD-Bench exhibit large-scale variations, cluttered backgrounds, and different seasons. Unlike existing datasets, RSSOD-Bench offers uniform distribution across scene categories. We benchmark \textbf{23} different state-of-the-art approaches from both the computer vision and remote sensing communities. Experimental results demonstrate that more research efforts are required for the RSSOD task.
\end{abstract}

\begin{keywords}
benchmark, dataset, remote sensing, salient object detection
\end{keywords}

\section{Introduction}
\label{sec:intro}
% building models and piecewise planarity
Automatically extracting salient objects from images can serve as an important pre-processing step for numerous computer vision and remote sensing tasks \cite{HFANet2022,SRAL2023, xiong2022earthnets, ABNet2022}. To name a few, salient object detection (SOD) has been applied in self-supervised learning \cite{van2021unsupervised}, image quality assessment \cite{gu2016saliency}, image retrieval \cite{babenko2015aggregating}, etc. SOD aims to extract visually distinctive objects from diverse complicated backgrounds at a pixel level. Namely, given an input image, two steps are required for SOD models to successfully detect the salient objects: 1) determine correct salient areas from cluttered backgrounds; 2) accurately segment the pixels of salient objects. 
\begin{figure*}[htp]
    \centering
    \includegraphics[width=1\linewidth]{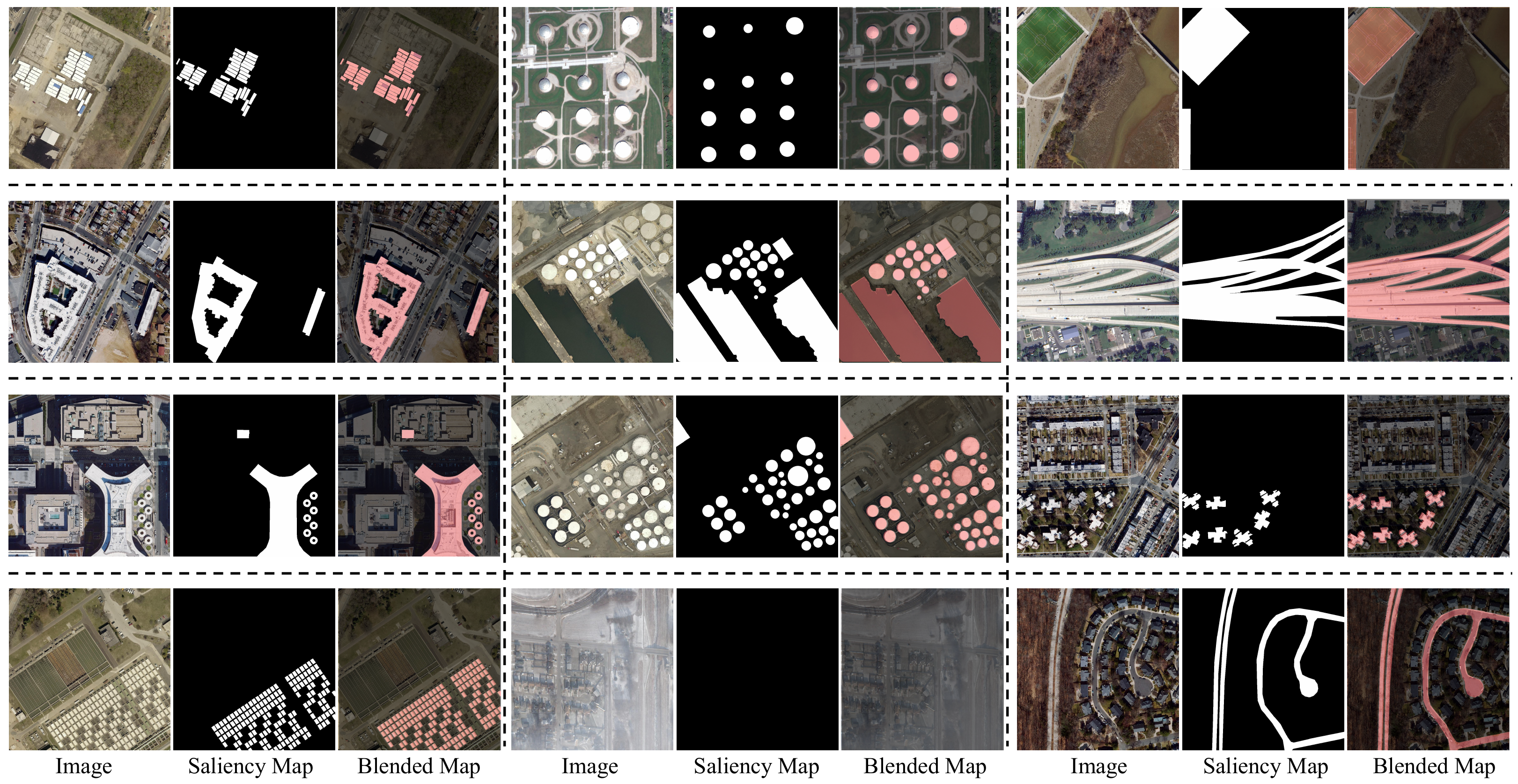}
    \vspace{-0.4cm}
    \caption{Illustration of some visualization examples in the RSSOD-Bench dataset.
        Several challenging examples are presented, including tiny, large, dense, incomplete, and irregular salient objects.
    }
    \label{fig:motivation}
\end{figure*}
% Difference between natural scene image and existing SOD datasets

For natural scene images, SOD has achieved remarkable success with the advent of deep learning. However, research on SOD for remote sensing and Earth observation data (RSSOD) is still in its infancy. Natural scene images are usually object-centric. In contrast, remote sensing imagery, captured through high-altitude shooting, usually covers a larger range of scenes with diverse ground objects and complicated backgrounds. Considering these differences, several datasets dedicated to remote sensing data have been released to foster the research of novel methods. The ORSSD dataset \cite{LVNet} contains 800 images (600 for training and 200 for testing) collected from Google Earth. EORSSD \cite{DAFNet} is an extended version of ORSSD with 2,000 images (1400 for training, 600 for testing) in total. ORSI-4199 \cite{MJRBM2021} is a large dataset, which contains 4,199 images with pixel-level annotations. 
\begin{table}
  \caption{Statistics comparison with existing RSSOD datasets. }
		\label{T1}
  \scalebox{0.78}{
			\begin{tabular}{c|c|c|c|c|c}
				\Xhline{1pt} 
				Datasets              & \#Total Images           & \#Cities          & \#Training          & \#Validation        & \#Test           \\ \hline
				ORSSD      & 800             & ---            & 600            & 0                  & 200                    \\
				EORSSD       & 2,000           & ---            & 1,400            & 0                   & 600           \\
				ORSI-4199      & 4,199             & ---            &  2,000            & 0                & 2,199          \\ \hline
				\textbf{Ours} & \textbf{6,000} & \textbf{4} & \textbf{3,000} & \textbf{600} & \textbf{2,400}     \\\Xhline{1pt}
		\end{tabular}}
  \vspace{-0.5cm}
\end{table}
\begin{table}[t]
	\caption{Comparison study with state-of-the-art methods on the proposed RSSOD-Bench dataset.}
	\centering
	\renewcommand{\arraystretch}{0.992}
	\resizebox{\linewidth}{!}{
	\begin{tabular}{l|c|c|c|c|c}
		\Xhline{1pt}
		Methods & Publications &$\text{MAE}\downarrow$ &$F_{\beta} \uparrow$    & $S_{m}\uparrow$ & $E_{m} \uparrow$\\\hline
		LC \cite{LC2006}&MM’06 &0.1834  &0.2467 & 0.5258 & 0.4826 \\
		FT\cite{FT2009} &CVPR’09 &0.1588 &0.2586 & 0.5417 &0.4859 \\\hline
		DSS\cite{HouPami19Dss} &CVPR’17 &0.0622 &0.6761 &0.7627 &0.7022 \\
		NLDF\cite{NLDF2017} &CVPR’17 &0.0535 &0.6765 &0.7798 &0.7632 \\
		RAS\cite{RASv12018} &ECCV’18 &0.0476 &0.7158 &0.7923 &0.7524\\
		PoolNet\cite{PoolNet2019} &CVPR’19 &0.0609&0.6915&0.7684&0.7092 \\
		PFAN\cite{PFAN2019} &CVPR’19 &0.0430 &0.7207 &0.8046&0.7684\\
		SCRN\cite{SCRN2019} &ICCV’19 &0.0415 &0.7435 &0.8117 &0.7582\\
		F3Net\cite{F3Net2020} &AAAI’20 &0.0456 &0.7089 &0.8043 &0.7679 \\
		MINet\cite{MINet2020} &CVPR’20 &0.0396 &0.7121 &0.8008 &0.8049\\
		GateNet\cite{GateNet2020} &ECCV’20 &0.0385 &0.7369 &0.8224  &0.7877 \\
		PFSNet\cite{PFSNet2021} &AAAI’21 &0.0409 & 0.7457 &0.8271 &0.8027 \\\hline
		SARNet\cite{SARNet2021} & RS’21 &0.0402&0.7298&0.8242&0.8165 \\
		DAFNet\cite{DAFNet} & TIP’21 &0.0541&0.7006&0.7939&0.7389 \\
		FSMINet\cite{FSMINet2022} &GRSL’22 &0.0406&0.7388&0.8186&0.8113\\
		MSCNet\cite{MSCNet2022} &ICPR’22 &0.0444 &0.7486 & 0.8102 &0.8016\\
		MCCNet\cite{MCCNet2021} &TGRS’22 &0.0417 &0.7525 &0.8287 &0.8122  \\
		CorrNet\cite{CoorNet2022} &TGRS’22 &0.0449 & 0.7415  &0.7914 &0.7503 \\
		MJRBM-R\cite{MJRBM2021} &TGRS’22 &\textcolor{blue}{0.0378} &0.7531 &0.8313 &0.7899 \\
		EMFI-R\cite{EMFINet2021} &TGRS’22  &\textcolor{green}{0.0377} &\textcolor{red}{0.7765} &\textcolor{green}{0.8400} &\textcolor{red}{0.8244} \\
		HFANet-R\cite{HFANet2022} &TGRS’22 &0.0393 &\textcolor{green}{0.7635} &0.8277 &0.8020 \\
		ACCo-V\cite{ACCoNet2022} &TCYB’23 &\textcolor{red}{0.0367} &\textcolor{blue}{0.7630} &\textcolor{red}{0.8406} &\textcolor{blue}{0.8225}\\
		ACCo-R\cite{ACCoNet2022} &TCYB’23 &0.0385 &0.7583 &\textcolor{blue}{0.8347} &\textcolor{green}{0.8226}\\%\hline
		%Ours &-- & & & & \\
		\Xhline{1pt}
	\end{tabular}
	\centering
	\label{table_2}
	\vspace{-1cm}
	}
\end{table}

Several shortcomings of existing RSSOD datasets hinder further research and progress of the SOD. The first limitation is that the scale of existing datasets is relatively small. As presented in Table \ref{T1}, we list the statistical information of different existing datasets. The number of images in ORSSD and EORSSD datasets is less than 2,000, which is not enough to align the performance of models well with real-world scenarios. In contrast, our dataset contains 6,000 images collected in four different cities, which is larger than the existing ones.

The second limitation is that the images are limited to some specific scene categories. The remote sensing images of existing datasets are usually collected in several scene categories and not uniformly sampled from the Earth's surface. This makes the data distribution does not align well with real-world applications. Considering this problem, we introduce RSSOD-Bench to facilitate the research in the community. %definition of SOD

\section{Dataset Construction}
For the RSSOD-Bench dataset, we annotate the salient objects that are naturally distinct from the background and are associated with certain object categories useful for specific tasks. Specifically, the following objects are annotated: buildings, lakes, rivers, highways, bridges, aircraft, ships, athletic fields, badminton courts, volleyball courts, baseball fields, basketball courts, gymnasiums, storage tanks, etc. As presented in Fig. \ref{fig:motivation}, the salient objects in RSSOD-Bench have large scale variations with tiny and large regions. Also, there are severely cluttered backgrounds with different seasons. In some scenes, the salient objects can be very dense. Note that, the remote sensing images in the RSSOD-Bench dataset are uniformly distributed regarding scene categories. This is different from existing ones that are usually collected from several scene categories.

% Benchmark results and challenges

\section{Methods and Results}
\label{sec:method}
We report four commonly-used metrics in the field of SOD, including the \textbf{MAE}~\cite{MAE}, \textbf{F-Measure}~\cite{FT2009}, \textbf{S-Measure}~\cite{Smeasure}, and \textbf{E-Measure}~\cite{Emeasure}. \textbf{MAE} quantifies the pixel-level disparity between the predicted saliency map (SM) and the ground truth (GT). \textbf{F-Measure} ($F_{\beta}$) is a composite metric that combines precision and recall to assess the similarity between SM and GT, with different weights. \textbf{S-Measure} ($S_{m}$) utilizes balanced structural information from object-aware and region-aware levels to evaluate the structural likeness between SM and GT. \textbf{E-Measure} ($E_{m}$) is a metric that signifies the degree of pixel-level correspondence and image-level statistics.

To comprehensively validate existing state-of-the-art methods on our proposed dataset, we conduct extensive experiments to benchmark and compare their performance. Specifically, LC \cite{LC2006} and FT\cite{FT2009} are classical saliency detection methods with no use of deep learning models. As there are considerable SOD methods proposed in the computer vision (CV) community, we choose ten typical deep learning-based methods, including DSS\cite{HouPami19Dss}, NLDF\cite{NLDF2017}, RAS\cite{RASv12018}, PoolNet\cite{PoolNet2019}, and so forth. Compared with classical methods, deep learning methods from the CV community can obtain clearly better results. For example, GateNet\cite{GateNet2020} and PFSNet\cite{PFSNet2021} can achieve a $S_{m}$ of over 0.82. Furthermore, 11 RSSOD methods are compared on our dataset, including the SARNet\cite{SARNet2021}, FSMINet\cite{FSMINet2022},	MCCNet\cite{MCCNet2021}, and so on. The methods from the remote sensing community achieve state-of-the-art performance. 

As presented in Table \ref{table_2}, we use different colors (i.e., red, green, and blue) to highlight the best, second-best, and third-best quantitative results. Typically, several latest algorithms, MJRBM-R\cite{MJRBM2021}, EMFI-R\cite{EMFINet2021}, HFANet-R\cite{HFANet2022}, ACCo-V\cite{ACCoNet2022}, and ACCo-R\cite{ACCoNet2022} works well on the proposed RSSOD-Bench dataset. This is in line with the expectations of these methods, and also reflects the feasibility of the proposed dataset. 

\begin{figure*}[htbp]
    \centering
    \includegraphics[scale=0.23]{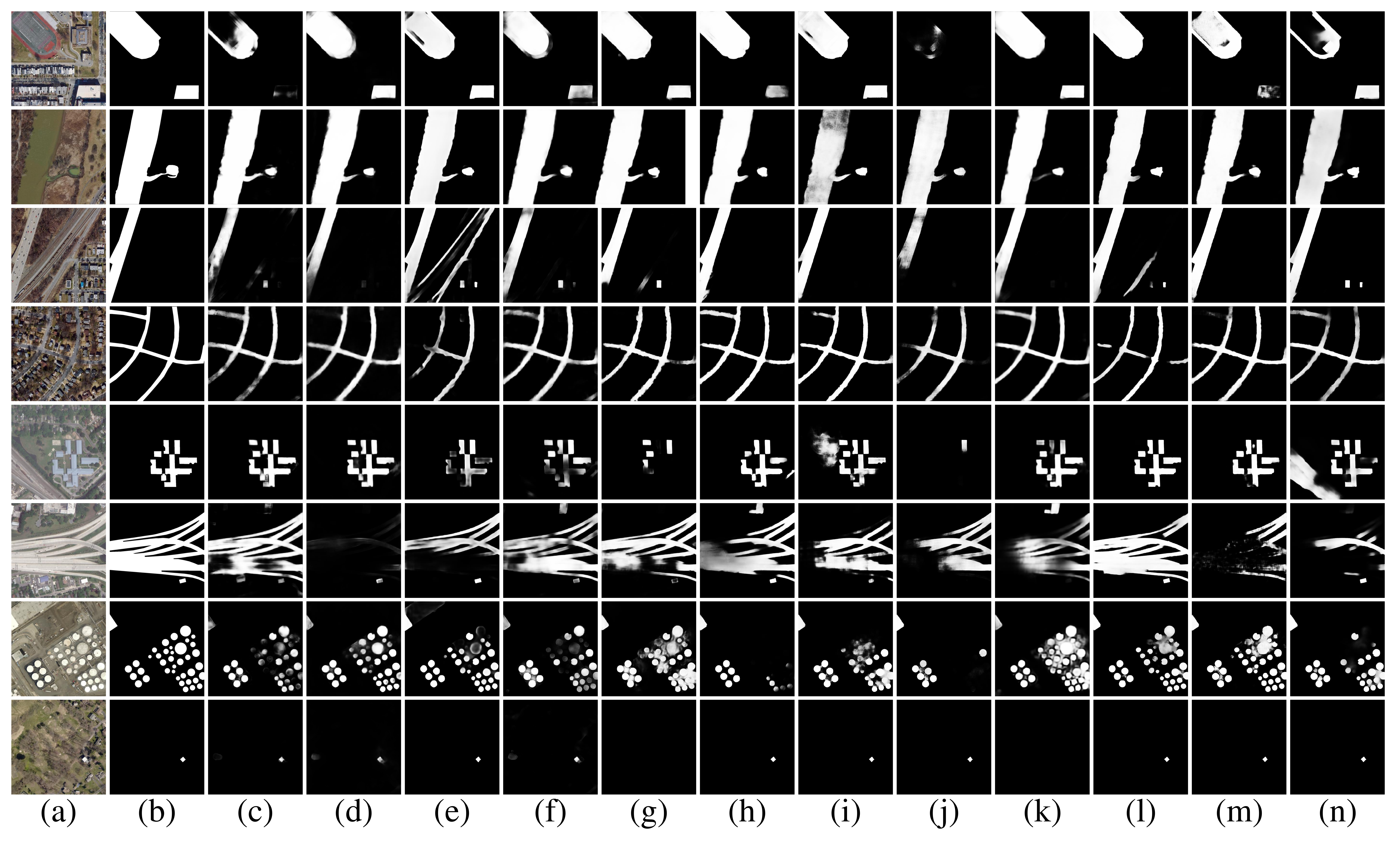}
    \vspace{-0.6cm}
    \caption{Visualization saliency maps with 12 state-of-the-art methods on the proposed dataset, including five NSI-SOD approaches, and seven RSI-SOD algorithms, on different patterns. (a) Optical RSIs. (b) GT. (c) PFAN. (d) SCRN. (e) F3Net. (f) GateNet. (g) PFSNet. (h) FSMINet. (i) MCCNet. (j) CorrNet. (k) MJRBM-R. (l) EMFINet-R. (m) HFANet-R. (n) ACCo-V.
    }
    \label{fig2}
\end{figure*}

However, as visualized in Fig. \ref{fig2}, the segmentation results of hard examples are still not satisfactory on the proposed dataset. This indicates that more research efforts are required to enhance the SOD performance for real-world applications.

\section{Conclusion}
We introduce the RSSOD-Bench dataset for salient object detection (SOD) in remote sensing imagery. Addressing the limitations of existing RSSOD datasets, RSSOD-Bench comprises carefully chosen images from four US cities, exhibiting diverse salient objects, varied backgrounds, and seasonal variations. Unlike previous datasets, RSSOD-Bench ensures uniform scene category distribution. We evaluate \textbf{23} state-of-the-art approaches from computer vision and remote sensing communities. While these methods perform well on RSSOD-Bench, there is still room for improving SOD accuracy compared to existing datasets. Therefore, further research efforts and advanced models are needed to enhance performance.

\small
%\bibliographystyle{IEEEbib}
%\bibliography{refs}
\bibliographystyle{IEEEtran}
\bibliography{IEEEfull,refs}
\end{document}